# Hybrid Model using Feature Extraction and Non-linear SVM for Brain Tumor Classification

Lalita Mishra[1], Shekhar Verma[1], Shirshu Varma[1]

[1] Indian Institute of Information Technology Allahabad,
Uttar Pradesh-211015, India
rsi2018502@iiita.ac.in

**Abstract.** It is essential to classify brain tumors from magnetic resonance imaging (MRI) accurately for better and timely treatment of the patients. In this paper, we propose a hybrid model, using VGG along with Nonlinear-SVM (Soft and Hard) to classify the brain tumors: glioma and pituitary and tumorous and non-tumorous. The VGG-SVM model is trained for two different datasets of two classes; thus, we perform binary classification. The VGG models are trained via the PyTorch python library to obtain the highest testing accuracy of tumor classification. The method is threefold, in the first step, we normalize and resize the images, and the second step consists of feature extraction through variants of the VGG model. The third step classified brain tumors using non-linear SVM (soft and hard). We have obtained 98.18% accuracy for the first dataset and 99.78% for the second dataset using VGG19. The classification accuracies for non-linear SVM are 95.50% and 97.98% with linear and rbf kernel and 97.95% for soft SVM with RBF kernel with D1 and 96.75% and 98.60% with linear and RBF kernel and 98.38% for soft SVM with RBF kernel with D2. Results indicate that the hybrid VGG-SVM model, especially VGG 19 with SVM, is able to outperform existing techniques and achieve high accuracy.

## 1 Introduction

Brain tumor (BT) is one of the deadly deceases which spread into the human body because of the growth of abnormal cells in the brain. It is the cancerous or noncancerous mass that grows abnormally in the brain. Cancer starts forming in other body parts except for the brain; however, it can spread to the brain. Further, the tumors can start growing in the brain and can be cancerous or noncancerous. According to the national center for biotechnology information (NCBI) 5 to 10 people in every 100000 population were registered in 2016, showing an increasing trend in tumor detection [1]. Dasgupta et al. [1] categorized five different types of frequent tumors were medulloblastoma (11.4%), ependymal tumors (4.8%), craniopharyngioma (9.7%), astrocytoma (47.3%), and nerve sheath tumors (4.1\%). Although, some of them are low incensed and can be treated if found in the initial stages. Thus, it is crucial to detect the tumor in its initial stages so one can take preventive measures for treatment. Nowadays, machine learning is one of the novel approaches to tumor detection. Moreover, we required to classify the tumor according to the international classification of diseases for ontology, which was defined by WHO [2]. Thus, classification has become an essential aspect of tumor detection. MRI image is one of the most accurate ways for tumor detection, and we can use machine learning techniques for image classification to detect the tumor. Most of the organization uses the grading-based method to categorize each tumor based upon its intensity or aggressiveness, as.

- **Grade 1** is generally a benign tumor, mostly found in children, and this type of tumor is mostly curable.
- **Grade 2** includes Astrocytomas, Oligodendrogliomas, and Oligoastrocytoma, commonly found in adults. It can progress slowly to high-grade tumors.
- **Grade 3** is generally the combination of Anaplastic Astrocytomas, Anaplastic Oligodendrogliomas or Anaplastic Oligoastrocytoma. These are quite aggressive and dangerous.
- **Grade 4** is generally glioma multiform tumor, the most aggressive tumor in the WHO category.

Machine learning can classify the image based on the characteristics of the MR image. So, we recognize three types of tissues for the tumor classification as: Tumor Core, the region of tumor tissues, which has the malignant cells that are actively proliferating. Necrosis identifies the region where tissues/cells are dying or dead. This characteristic is used to differentiate between low grades of gliomas and GBM and perifocal oedema occurs because of the glial cell distribution in the brain. Because of this, a swelled region forms around the brain, and fluid is filled around the tumor core.

Imaging methods such as MRI can be used for diagnosis. MRI image [3] helps us to identify the location where the tumor formed. MR images provide the physiology, anatomy of the lesion, and metabolic activity with its haemodynamic. Thus, MR images are used for primary diagnosis. However, errors in the manual detection with these methods can be threatening to life. The detection accuracy of the tumor is significant; manual processes are time-consuming and unreliable. Therefore we need appropriate precise detection and classification measures for BTs.

Non-Linear SVMs are utilized to classify non-linearly separable data that is, if, for any dataset, the classification cannot be performed using straight line, then it is non-linear data, and to classify such data, non-linear classifiers are used, such as non-linear SVM classifier. However, SVM either takes each image as a data points or requires handcrafted features. Both limit the utility and accuracy of SVM. CNN is able to extract features with manual intervention. However, accuracy of these methods needs to be improved.

*Novelty and Contribution of the study:*

We have proposed a two-stage model of BT classification using the VGG model with SVM, VGG-SVM, to acquire performance enhancement. In other studies, researchers have applied these models individually and obtained good results; however, the integration of various SVMs with VGG models results validate our approach. Our major contributions can be listed as:
- We have utilized untrained VGG models (VGG11, VGG13, VGG16, and VGG19)for feature extraction and trained the models with MRI images containing brain tumors.
- We have flattened the extracted features and passed these extracted features to non-linear SVMs (soft and hard) for classification brain tumor classification.

## 2 Related Works

In this section, we discuss the existing work and compare the existing classification results. We have covered several papers utilizing VGG models, variants of SVM and some integrated models for BT classification and compare their performance in **Table 1**, which combines the working model and the obtained results by each author.

Khan et al. [4] applied k-means and discrete cosine transform algorithm to partition the dataset, transfer learning for feature extraction, and ELM for tumor classification and achieved 97.90% classification accuracy for a BT. Sajjad et al. in [5] segmented the tumor using InputCascadeCNN, having two streams, one is $7 \times 7$ fields for local feature extraction, and the other is $13 \times 13$ field for global feature extraction. They used rotation, skewness, flipping, and shears for extensive data augmentation and transfer learning approach for pattern learning and VGG19 for tumor grade classification. Latif et al. in [6] utilized a 17-layer CNN for feature extraction and, after that, used several machine learning classifiers, like, SVM, MLP, NB, and RF, for tumor classification. They have obtained the best classification results using SVM. Whereas in [7] by Senan et al., minmax normalization was applied before feature extraction and then classified the tumor grade by AlexNet, Resnet18, AlexNet with SVM, and ResNet18 with SVM models. Vadhnani et al. in [8] utilized Otsu thresholding for tumor segmentation; discrete wavelet transforms algorithm for feature extraction, principal component analysis for reducing the extracted features, and SVM for tumor type classification.

Bodapati et al. in [9] first used k-means clustering for tumor segmentation from MR images and then applied a median filter on the segmented tumor data. They measured the distance from each cluster center using the Euclidean distance and classified the BT using improved SVM. Rajinikanth et al. customize the pre-trained VGG19 model by using serially fused deep features and handcrafted features and then classify the BT using several classifiers, including VGG19, linear SVM and kernel SVM in [10]. Tazin et al. in [11] used MobileNetV2 for image preprocessing, transfer learning to categorize data, and VGG19 for tumor classification. N. Nar et al. in [12] applied rotation and shifting operations together with scaling for data augmentation. After the augmentation, they applied transfer learning for feature extraction from MR images and VGG19 along with several DL models for tumor classification.

We have listed the obtained classification accuracy by each surveyed paper in **Table 1** to provide a comparative view of the results. Although they have obtained good results, however, we have obtained better classification accuracy using a variety of VGG models in integration with non-linear SVM for BTs in VGG-SVM.

Table 1 Comparative Literature Survey

| Paper | Model Used | Classification Accuracy |
|---|---|---|
| Khan et al. [4] 2020 | VGG19 + ELM | 97.90% |
| Sajjad et al. [5] 2019 | VGG19 | 94.58% |
| Latif et al. [6] 2022 | SVM | 96.19% for HGG<br>95.46% for LGG |
| Senan et al. [7] 2022 | AlexNet + SVM,<br>ResNet18 + SVM | 95.10%<br>91.20% |
| Vadhnani et al. [8] 2022 | SVM(RBF),<br>SVM(Linear) | 97.6%<br>94% |
| Bodapati et al. [9] 2022 | K-means + ISVM | 95% |
| Rajnikanth et al. [10] 2020 | VGG19,<br>SVM(RBF),<br>SVM(Linear) | 95.90%,<br>95.30%,<br>95.60% |
| Tazin et al. [11] 2021 | VGG19 | 88.22% |
| Nar et al. [12] 2022 | VGG19 | 97.20% |

We have reviewed several recent approaches, as shown in **Table 1**, and compared them in terms of their model and classification accuracy. It is illustrated in **Table 1** that these state-of-the-art works lack in providing better tumor classification. Thus, we propose an integrated model for tumor classification.

## 3 VGG-SVM Model

In this section, we present VGG-SVM architectural and implementation details along with the description of the evaluation datasets. It describes the flow and working of the model. It consists of three parts, as follows:

**3.1 Model Architecture**

In this work, we integrate the VGG model with non-linear SVM.

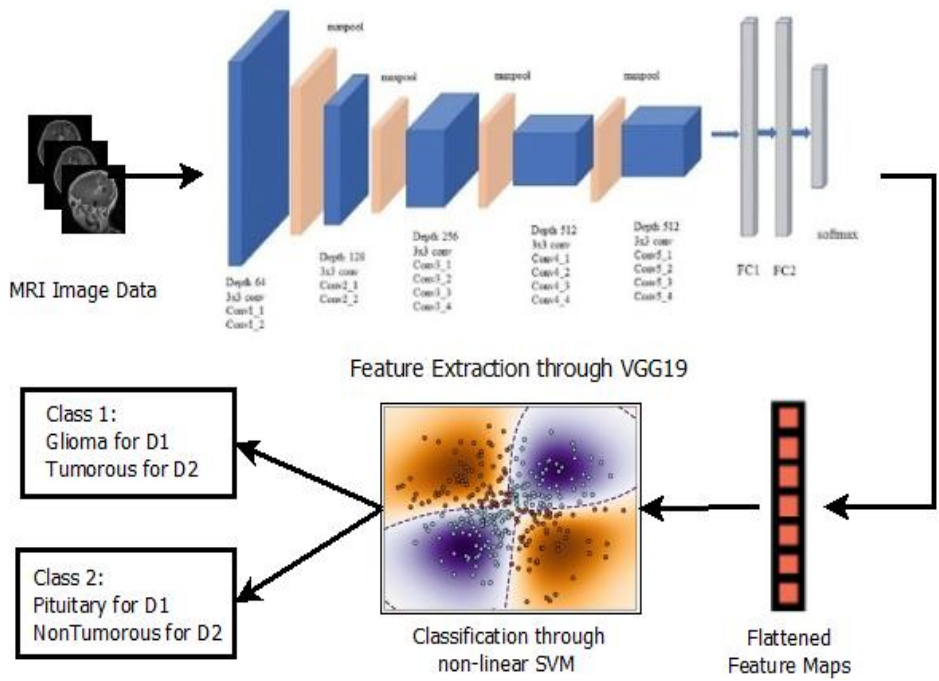

Figure 1 Feature extraction and Classification Model

In Figure 1 the architecture of the VGG-SVM model is shown. Here, we have used VGG19 architecture only. Variation of VGG models is applied similarly, with several non-linear SVM variants.

*Feature Extraction:*
Variations of VGG models are used to extract the MRI features. Features are defined as important information about a particular task. For MRI image processing, features are the texture, intensity value, shape, and boundary of a particular image. Large datasets contain an enormous amount of variables, among which only some are important for that particular task. Feature extraction is the segregation of the most important information.

In our case, extracted features are the tumor characteristics, such as tumor area, circularity, tumor shape, etc. The features are extracted using the fully connected layer of VGG models. We have selected the VGG model for feature extraction because of its deep nature and usage of the most minor receptive field, that is, $3 \times 3$, and stride 1. Thus, they provide fewer parameters & the most relevant features at the fully connected layer compared to other DL models having $11 \times 11$ or $7 \times 7$ receptive fields with 4, 5, and higher stride values. Less number of parameters provides fast convergence of the model and reduces the overfitting problem. Among the variety of VGG models, VGG16 and VGG19 provide a significant change in performance.

*Classification:*
These extracted features are then fed to non-linear SVM for classifying the tumor type. We have used non-linear SVM because, in our case, we have non-linear data of BTs in the form of MR images. We have applied SVM with soft and hard margins to cover each possibility for improving the classification accuracy.

We have provided the extracted features from each VGG model to the non-linear SVM with RBF and linear kernel and non-linear SVM with soft margins to obtain the final tumor classification.

### 3.2 Implementation Details and Preprocessing

In this paper, we applied the binary classification of BTs among glioma & pituitary tumors and tumorous & non-tumorous images. Before classification, we have extracted the image features using VGG models.

*Feature Extraction:*
We resize the images as $224 \times 224$. Then, we have created *X train; y train; X test*, and *y test* using 7 : 3 random split for training and validation sets for all VGG models. We have used the batch size of 32 at the time of training the VGG model, with a 0.001 learning rate. We have trained the VGG models for 200 epochs. After training the VGG model, we changed the batch size to 1 at a time, extracted the image features from the third last fully connected layer to obtain the maximum number of extracted features and minimize the loss, and stored these extracted features in the form of NumPy array.

*Classification:*
We then pass the extracted features through VGG models to non-linear SVM for binary classification of the BT MRI. We have used RBF and linear kernel functions with the hard SVM and RBF kernel with the soft SVM. VGG and non-linear SVM models are implemented using PyTorch, with a python script, both with Google Colab. The VGG and SVM hyperparameters are illustrated in **Table 2**.

Table 2 Model Hyperparameters

| Hyperparameters | Value |
|---|---|
| Optimizer | Optim |
| Image Size | 224 × 224 |
| Weight | VGG19 |
| Loss | Crossentropy |
| Matrices | Accuracy and Loss (VGG)<br>Accuracy, Precision, Recall, f1-score, Hinge Loss (SVM) |
| Epochs | 200 |
| Batch Size | 32 (VGG Training)<br>1 (Feature Extraction) |
| Activation Function | ReLU |
| Learning Rate | 0.001 |
| Error Control Parameter(C) | 0.001 |
| Gamma ($\gamma$) | 0.001 |

### 3.3 Dataset Description

We have used two different datasets for classification and better comparative results. We have obtained all the five datasets from Kaggle and all the datasets are publically available and containing MRI images. The first dataset (D1) [13] contains total of 1800 images (900 Glioma and 900 Pituitary). The second dataset (D2) consists of 3000 images (1500 Tumorous and 1500 Non-tumorous) downloaded from [14].

Table 3 Dataset Description

| Dataset | Total No. of Images | Image Type |
|---|---|---|
| D1 | 1800 | Glioma (900) and Pituitary (900) |
| D2 | 3000 | 1500 Tumorous and 1500 Non-tumorous |

## 4 Experimental Results and Discussion

### 4.1 Results

We have used accuracy and loss evaluation metrics for VGG models and accuracy, precision, recall, f1-score, and hinge loss for non-linear SVMs. The mathematical formulation of these evaluation metrics is given using following equations:

$$Accuracy = \frac{(TP + TN)}{(TP + TN + FP + FN)}$$

$$Precision = \frac{TP}{(TP + FP)}$$

$$Senstivity = \frac{TP}{(TP + FN)}$$

$$F - score = \frac{2TP}{(2TP + FP + FN)}$$

$$HingeLoss = \max\{0, 1 - y_i(\omega^T x_i + b_i)\}$$

where, TP is true positive, TN is true negative, FP is false positive, and FN is false negative, $\omega$ and $b$ are parameters of hyperplane and $x_i$ is input variable.

We have shown the graphical representation of obtained accuracies in a comparative way in Figure 2, using all VGG models during model training for D1. Figure 2 shows that the obtained accuracies are showing much difference in initial epochs, however, with each increasing epochs the difference converges. The final classification results obtained from non-linear SVMs are illustrated in Tables 3 and 4 for the first and second datasets, respectively. We have shown all the results, including VGG training and classification using non-Linear SVM with linear kernel, non-Linear SVM with RBF kernel, and non-Linear SVM with soft margins and RBF kernel in the Tables 3 and 4. Further, the confusion matrix for D1 and D2 is shown in Fig. 3 for non-linear SVM with RBF kernel.

Table 4 Our Experimental results for First Dataset

| | **VGG11** | **VGG13** | **VGG16** | **VGG19** |
|---|---|---|---|---|
| **VGG Training** | | | | |
| **Accuracy** | 91.37% | 95.76% | 96.75% | **98.18%** |
| **Loss** | 0.0076 | 0.0045 | **0.0022** | 0.0031 |
| **Non-Linear SVM with Linear Kernel** | | | | |
| **Accuracy** | 92.56% | 94.19% | **96.20%** | 95.50% |
| **Precision** | 0.8760 | 0.9332 | 0.9250 | 0.9440 |
| **Sensitivity** | 0.9751 | 0.9623 | 0.9250 | 0.9733 |
| **F - score** | 0.9223 | 0.9475 | 0.9250 | 0.9584 |
| **Hinge Loss** | 0.098 | 0.079 | **0.0382** | 0.0395 |
| **Non-Linear SVM with RBF Kernel** | | | | |
| **Accuracy** | 93.06% | 94.89% | 96.79% | **97.98%** |
| **Precision** | 0.9702 | 0.9439 | 0.9650 | 0.9798 |
| **Sensitivity** | 0.9702 | 0.9439 | 0.9650 | 1 |
| **F - score** | 0.9702 | 0.9439 | 0.9650 | 0.9798 |
| **Hinge Loss** | 0.096 | 0.081 | **0.0376** | 0.0391 |
| **SVM with soft margin and RBF Kernel** | | | | |
| **Accuracy** | 93.45% | 94.19% | 95.91% | **98.76%** |
| **Precision** | 0.9345 | 0.9419 | 0.9591 | 0.9795 |
| **Sensitivity** | 0.9347 | 0.9474 | 0.9641 | 1 |
| **F - score** | 0.9345 | 0.9446 | 0.9615 | 0.9896 |
| **Hinge Loss** | 0.1192 | 0.0985 | **0.0235** | 0.0398 |

### 4.2 Discussion

We have obtained the best classification accuracy with non-Linear SVM with soft margins and RBF kernel after the hyperparameter tuning with the output of the VGG19 model; however, the loss values are minimum with the VGG16 model. By referring to Table 1 from Section 2, we obtain that the surveyed models did not achieve the classification accuracy as VGG-SVM. Thus, VGGSVM outperforms the state-of-the-art models using VGG19 and non-linear SVM with soft margins and RBF kernel and other respective variants. We have shown the best values of accuracy and loss for each model in bold letters to make it easy to compare, for each model, in

Tables 3 and 4, and found that the best result for classification accuracy is obtained after training with VGG19 and classify using non-linear SVM with soft margins and for loss is obtained after training with VGG16 and non-linear SVM with soft margins, both for D2.

Table 5 Our Experimental results for Second Dataset

|  | VGG11 | VGG13 | VGG16 | VGG19 |
|---|---|---|---|---|
| VGG Training | | | | |
| Accuracy | 92.95% | 95.91% | 99.12% | **99.78%** |
| Loss | 0.0079 | 0.0046 | **0.0013** | 0.0019 |
| **Non-Linear SVM with Linear Kernel** | | | | |
| Accuracy | 93.51% | 95.23% | **96.83%** | 96.75% |
| Precision | 0.8910 | 0.9389 | 0.9625 | 0.9610 |
| Sensitivity | 0.8910 | 0.9389 | 0.9650 | 0.9610 |
| F - score | 0.8910 | 0.9389 | 0.9637 | 0.9610 |
| Hinge Loss | 0.0923 | 0.0721 | **0.0265** | 0.0299 |
| **Non-Linear SVM with RBF Kernel** | | | | |
| Accuracy | 94.70% | 97.33% | 98.06% | **98.60%** |
| Precision | 0.9120 | 0.9733 | 0.9806 | 0.9841 |
| Sensitivity | 0.9020 | 1 | 1 | 0.9860 |
| F - score | 0.9069 | 0.9685 | 0.9792 | 0.9850 |
| Hinge Loss | 0.087 | 0.0532 | **0.0387** | 0.0391 |
| **SVM with soft margin and RBF Kernel** | | | | |
| Accuracy | 94.35% | 95.16% | 96.37% | **99.85%** |
| Precision | 0.9435 | 0.9516 | 0.9637 | 0.9838 |
| Sensitivity | 0.9564 | 0.9865 | 1 | 1 |
| F - score | 0.9499 | 0.9687 | 0.9815 | 0.9918 |
| Hinge Loss | 0.1129 | 0.0964 | **0.0425** | 0.0492 |

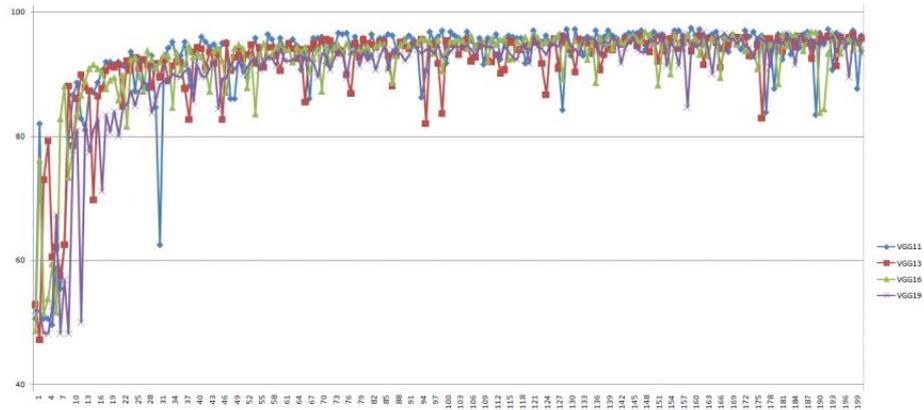

**Figure 2 Comparison of Accuracy using different VGG Models**

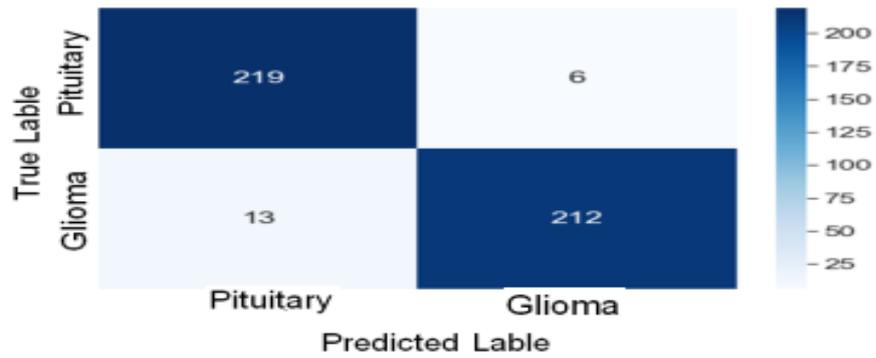

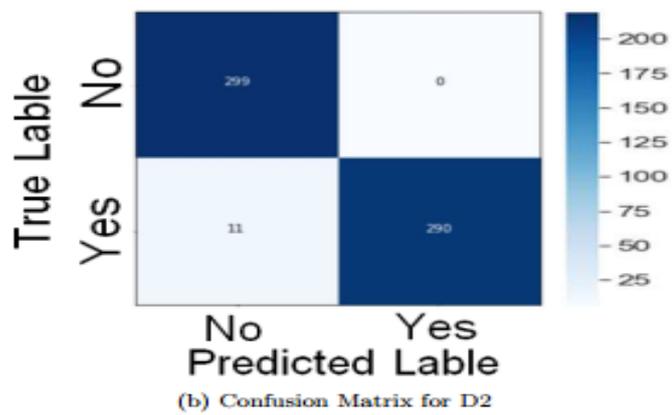

Figure 3 Confusion Matrix with RBF Kernel Non-linear SVM for D1 and D2

## 5  Conclusion

This paper provides an automatic and hybrid classification model for MRI based BT classification by integrating non-linear SVM with VGG models. Various VGG models (VGG11, VGG13, VGG16, and VGG19) are trained with MRI images of BTs (two datasets, D1 and D2) for feature extraction, and a variety of non-linear SVMs for classification (with linear and RBF kernel, and SVM with soft margins and RBF kernel), with the soft SVM with RBF kernel achieving the best performance (best classification accuracy with VGG19 (99.85%) and minimum loss with VGG16 (0.0425) for the second dataset. The VGG-SVM produced better classification accuracy than the existing models using Table 1. The highest accuracy achieved by existing models, integrating VGG19 with an extreme learning machine (ELM), was 97.90%. We have obtained 98.18% accuracy with the VGG19 model during training, 97.98% classification accuracy using non-linear SVM with RBF kernel, and 98.76% classification accuracy using non-linear soft SVM with RBF kernel for D1, and 99.78% accuracy with VGG19 model during training, 98.60% classification accuracy using non-linear SVM with RBF kernel, and 99.85% classification accuracy using non-linear soft SVM with RBF kernel, for D2. We can conclude that feature extraction through VGG and classification through SVM obviates the need of manual feature extraction and also yields higher accuracy as compared to standalone classifiers. However, this is possible with higher number of layers in VGG that gives better features along with soft margin non-linear SVM.